\documentclass[runningheads]{llncs}
\usepackage{graphicx}
\usepackage{subcaption}
\usepackage{amsmath,amssymb} 
\usepackage{color}
\usepackage[inline]{enumitem}

\usepackage[width=122mm,left=12mm,paperwidth=146mm,height=193mm,top=12mm,paperheight=217mm]{geometry}

\begin{document}

\pagestyle{headings}
\mainmatter
\def\ECCV18SubNumber{1635}  

\title{Hierarchical Video Understanding}

\authorrunning{Mahdisoltani et al.} 

\author{Farzaneh Mahdisoltani\textsuperscript{1,2}, Roland Memisevic\textsuperscript{2}, David Fleet\textsuperscript{1}\\
}
\institute{University of Toronto\textsuperscript{1}, Twenty Billion Neurons\textsuperscript{2} \\ 
\textsuperscript{1} \texttt{\{farzaneh, fleet\}@cs.toronto.edu}, 
\texttt{\{firstname.lastname\}@twentybn.com}
 }

\maketitle

\begin{abstract} 
We introduce a hierarchical architecture for video understanding that exploits the structure of real world actions by capturing targets at different levels of granularity. We design the model such that it first learns simpler coarse-grained tasks, and then moves on to learn more fine-grained targets. 
The model is trained with a joint loss on different granularity levels. We demonstrate empirical results on the recent release of Something-Something\footnote{Second release of Something-Something is used throughout this paper.} dataset, which provides a hierarchy of targets, namely coarse-grained action groups, fine-grained action categories, and captions. Experiments suggest that models that exploit targets at different levels of granularity achieve better performance on all levels.

\end{abstract}

\section{Introduction}
Actions in the real world are structured, as are objects. We design an architecture that exploits some semantics of this structure through a target hierarchy. In particular, we show networks extract better features by learning targets at different levels of granularity, yielding better performance at all levels.
We use Something-Something \cite{GoyalICCV17} dataset for our experiments. 
It comprises over $220,000$ videos and a hierarchy of targets, namely, 50 coarse-grained action groups, 
174 fine-grained action categories, and video captions with information about the actions, the objects, and their properties. Our contribution is two-fold: 
\begin{enumerate*}
    \item We present a hierarchical architecture that makes use of different levels of task granularity by
    jointly learning fine-grained and coarse-grained targets, as well as captions. 
    \item We show that such model perform better than networks trained solely on one level of target labels. 
\end{enumerate*}

\section{Related work} The literature on action recognition is vast (e.g.\ \cite{tran2015learning,LaptevCVPR08,carreira2017quo}), as is the growing literature 
on video captioning \cite{donahue2015long,venugopalan2014translating,tesselation}, this 
paper focuses on architecture to learn a common encoding for both recognition and 
captioning, leveraging different levels of granularity in video targets, which is not 
studied extensively to date.

\section{Approach}
Our deep neural network leverages information at different levels of granularity. It simultaneously performs coarse-grained classification of action groups, fine-grained classification of action categories, and video captioning.  It is trained end-to-end with a joint loss on all aforementioned targets. 

\begin{figure}[t!]
    \centering
    \includegraphics[width=13cm]{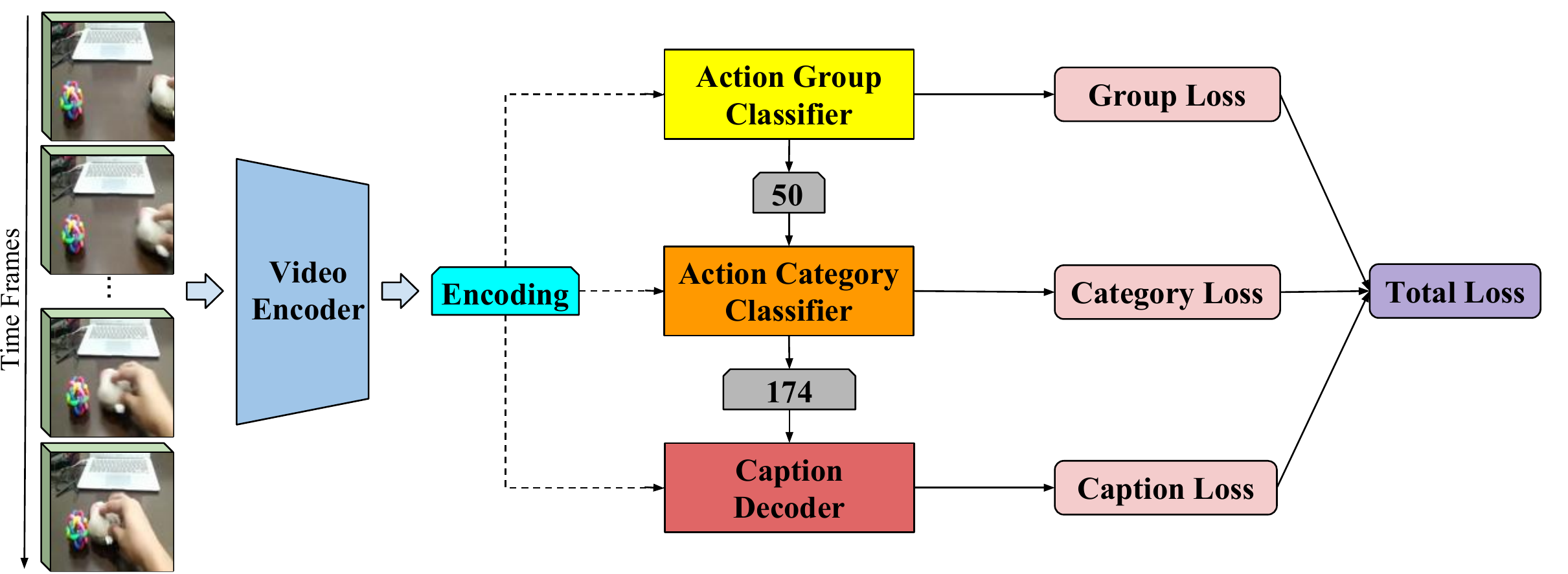}
    \caption{Our hierarchical architecture for targets with different levels of granularity, along with the loss functions on different levels.}
    \label{fig:arch}
    \vspace*{-6mm}
\end{figure}

The overall architecture is depicted in Fig.\ \ref{fig:arch}.
Video frames are fed into the video encoder, resulting in encoding vector $V$, which is then passed to processing modules (classifiers or captioners) for targets with different granularity.
Let $i$ denote the $i^{th}$ level of target granularity, with higher values of $i$ representing more fine-grained targets. Let $T_i$ be the set of targets at level $i$. In order to capture the structure 
between actions, the concatenation of probability distribution over $T_i$ and $V$ is passed 
to $(i+1)^{th}$ processing module. The model then tries to estimate the probability over $T_{i+1}$. 
In our model, $\forall c \in T_{i+1} : P(c|V) = \Sigma_{g \in T_{i}}{ P(c|g,V) P(g|V)}$.
In other words, at level $i+1$, given $p(g|V)$, instead of estimating the posterior $p(c|V)$ directly, we implicitly estimate $P(c|g,V)$, integrating out $g$. Decomposing the posterior in this way is expected to simplify learning, as it provides some hints from coarse- to fine-grained classification. The total loss being optimized, is sum of losses of different levels: $loss = \Sigma_{i} {w_i * loss_i}$. 

\begin{figure}[t!]
    \centering
    \includegraphics[width=13cm]{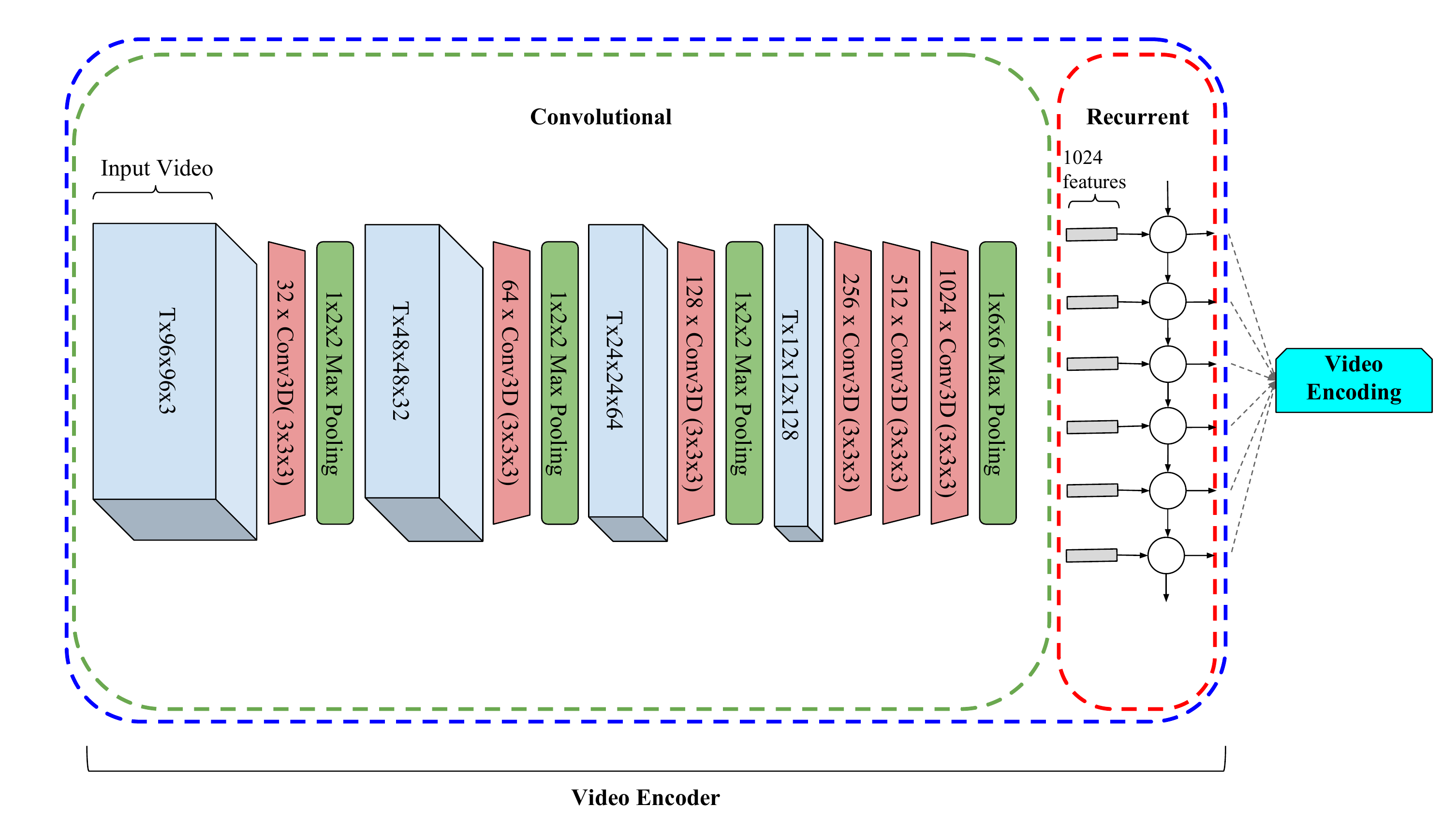}
    \caption{Our video encoder architecture includes 6 layers of 3DCNN, and 4 layers of spatial max-pooling followed by an LSTM layer. $T$ denotes the number of frames across time.}
    \label{fig:vid_encoder}
        \vspace*{-4mm}
\end{figure}

As mentioned before, Something-Something has 3 levels of targets.
In the first step (yellow box in Fig.\ \ref{fig:arch}), the model infers the action group, e.g. ``Putting'', vs. ``Taking''. In the second step  (orange box in Fig.\ \ref{fig:arch}), the output distribution over coarse-grained groups  (grey boxes in Fig.\ \ref{fig:arch}) is used along with video features to infer the fine-grained action category, e.g. if the model has figured out the group ``Putting'' in the previous step, here it needs to distinguish between ``Putting on top'', ``Putting behind'', and ``Putting next to''. 
We move one step further with more details, where the model needs to generate a caption (red box in Fig.\ \ref{fig:arch}), and describe what is happening in the video. At this step, the output distribution over fine-grained classes is passed to the caption decoder. For an accurate caption, the model needs to 
recognize and explain objects and their properties as well as the action. The rest of the section 
explains each component of our architecture in more details.\\
\textbf{Video Encoder.}
We use a minimal video encoder with 6 layers of 3D-CNN \cite{3DCNN} with $3 \times 3 \times 3$ filters, and 4 spatial max pooling layers (see Fig.\ \ref{fig:vid_encoder}). For temporal aggregation we use a bidirectional LSTM.  We then average the outputs over different time steps, yielding the final video encoding $V$.\\
\textbf{Video Classification.}
For classification at level $i$, we pass $V$ to a fully connected layer of size $|V|\times |T_i|$ followed by a softmax layer, $T_i$ being the set of targets at level $i$. Note that the dimensions of this layer varies for different levels $i$. 
We train classifiers using the usual cross-entropy loss over classes at the intended target level.\\
\textbf{Video Captioning.}
We use a two-layer LSTM for decoding captions. At each time-step, we use a softmax over the vocabulary words, conditioned on previously generated words. The captioning loss is the negative log-probability of the word sequence.

\section{Experiments} 
As a baseline, we train the network for each specific target level separately. To show the effectiveness of training on the target hierarchy, we first train the group classifier, then activate the category classifier, and finally add the captioner, see Table \ \ref{table:results}. The results reveal that by adding each level of detail, there is a boost in model performance, on all the target levels it's been trained on. 
Captioning accuracy measures the fraction of captions that the model got correct, word by word. Note that this is a very harsh metric, and the model does not get any score if it misses even one word of the correct caption. We are not investigating other metrics, as this experiment is not about that, and we only need to involve the captioning in the loss.


\textbf{Training settings.} We use frame rate of $12 fps$, and randomly pick $48$ consecutive 
frames. Videos with less than $48$ frames are padded by replicating  
first and last frames. We re-size frames to $128\times128$, and then use random cropping of 
size $96\times96$. For validation and testing, we use
center cropping with same size. We optimize all models using Adam, with an initial learning rate 
of $0.001$. Each step is finished after $20$ epochs of training over the full dataset.
\begin{table}[h!] 
\vspace*{-4mm}
\center
\begin{tabular}{l|c|c|c}
 {\bf\begin{tabular}[c]{@{}l@{}}Loss Weight\\ Vector\end{tabular}}&
 {\bf\begin{tabular}[c]{@{}l@{}}Coarse-grained\\ Accuracy\end{tabular}}&
 {\bf\begin{tabular}[c]{@{}l@{}}Fine-grained\\ Accuracy\end{tabular}}&
 {\bf\begin{tabular}[c]{@{}l@{}}Captioning \\Accuracy \end{tabular}} \\  \hline \hline
   $W=[0, 1, 0]$ & {0} & {48.40} & 0\\ \hline 
   $W=[0, 0, 1]$ & {0} & {0} & 3.13\\ \hline \hline
   $W=[1, 0, 0]$ & 58.04  & 0 & 0\\ \hline
   $W=[1, 1, 0]$ & {61.17} & {48.94} & 0\\ \hline
   $W=[1, 1, 1]$ & {61.65} & {50.28} & 3.34\\ \hline
\end{tabular}
\caption{Comparing models trained with different weight vectors for joint 
loss. $W$ shows weights on different level's losses, with the rightmost weight denoting the weight for captioning loss.}
\label{table:results}
\vspace*{-10mm}
\end{table}
\section{Conclusions}
We present a new architecture that leverages information from hierarchy of targets, at different levels of granularity. To train the architecture we use a joint loss on the different target levels.  This makes it possible to simultaneously classify the input video in different levels and produce a caption.  
Our  experiments show that the architecture achieves a performance boost over models that aim to solve 
the tasks individually. 

\bibliographystyle{splncs}
\bibliography{egbib}

\clearpage

\bibliographystyle{splncs}

\end{document}